%% file: main_cvpr21.tex
\documentclass[final]{cvpr}

\usepackage{times}
\usepackage{epsfig}
\usepackage{graphicx}
\usepackage{amsmath}
\usepackage{amssymb}

\usepackage[pagebackref=true,breaklinks=true,colorlinks,bookmarks=false]{hyperref}
\usepackage[square,numbers,sort&compress]{natbib}

\input{math_definition}

\input{imports}
\input{misc_commands}

\usepackage{hyperref}
\usepackage{url}

\newcommand{\ourmodel}{{\textsc{hammer}}\xspace}
\newcommand{\ourflat}{{\textsc{flat}}\xspace}

\newcommand{\SM}{{Suppl. Material}}

\newcommand{\footremember}[2]{%
   \thanks{\xspace\xspace#2}
    \newcounter{#1}
    \setcounter{#1}{\value{footnote}}%
}
\newcommand{\footrecall}[1]{%
    \footnotemark[\value{#1}]%
}

\begin{document}

%%%%%%%%% TITLE
\title{A Hierarchical Multi-Modal Encoder for Moment Localization in Video Corpus}
\author{
  Bowen Zhang$^{1}$\footremember{Google}{Work done while at Google} \footremember{Equal}{Authors Contributed Equally} \\
  {\tt\small zhan734@usc.edu} \and
  Hexiang Hu$^{1}$\footrecall{Google} \xspace\xspace \footrecall{Equal} \\
  {\tt\small hexiangh@usc.edu} \and
  Joonseok Lee$^{2}$\footrecall{Equal} \\
  {\tt\small joonseok@google.com} \and
  Ming Zhao$^{2}$\\
  {\tt\small astroming@google.com} \and
  Sheide Chammas$^{2}$\\
  {\tt\small sheide@google.com} \and
  Vihan Jain$^{2}$\\
  {\tt\small vihanjain@google.com} \and
  Eugene Ie$^{2}$\\
  {\tt\small eugeneie@google.com} \and
  Fei Sha$^{2}$\footremember{leave}{On leave from USC (feisha@usc.edu)} \\
  {\tt\small fsha@google.com}\and
  $^{1}$U. of Southern California\and
  $^{2}$Google Research}

\maketitle

%%%%%%%%% ABSTRACT
\begin{abstract}
Identifying a short segment in a long video that semantically matches a text query is a challenging task that has important application potentials in language-based video search, browsing, and navigation. Typical retrieval systems respond to a query with either a whole video or a pre-defined video segment, but it is challenging to localize undefined segments in untrimmed and unsegmented videos where exhaustively searching over all possible segments is intractable. The outstanding challenge is that the representation of a video must account for different levels of granularity in the temporal domain. To tackle this problem, we propose the HierArchical Multi-Modal EncodeR (HAMMER) that encodes a video at both the coarse-grained clip level and the fine-grained frame level to extract information at different scales based on multiple subtasks, namely, video retrieval, segment temporal localization, and masked language modeling. We conduct extensive experiments to evaluate our model on moment localization in video corpus on ActivityNet Captions and TVR datasets. Our approach outperforms the previous methods as well as strong baselines, establishing new state-of-the-art for this task.
\end{abstract}

\input{1_intro}
\input{2_related}
\input{3_method}
\input{4_exp}
\input{5_conclusion}

\section{Supplmentary Material}
\noindent In this section, we provide additional implementation details and visualization omitted in the main text.

\input{supp_content}
{\small
\bibliographystyle{ieee_fullname}
\bibliography{egbib}
}

\end{document}

%% file: math_definition.tex
\usepackage{amssymb}
\usepackage{amsmath,amsfonts}
\usepackage{amsopn,bm,mathtools}
\usepackage{bbm}

\usepackage{nicefrac}       % compact symbols for 1/2, etc.
\newcommand{\vct}[1]{\boldsymbol{#1}} % vector
\newcommand{\cst}[1]{\mathsf{#1}}  % constant

\newcommand{\ProbOpr}[1]{\mathbb{#1}}
\newcommand{\expect}[2]{%
\ifthenelse{\equal{#2}{}}{\ProbOpr{E}_{#1}}
{\ifthenelse{\equal{#1}{}}{\ProbOpr{E}\left[#2\right]}{\ProbOpr{E}_{#1}\left[#2\right]}}} % Expectation: syntax: E{1}{2} = E_1[2], E{}{2}=E[2], E{1}{} = E_1
\newcommand{\var}[2]{%
\ifthenelse{\equal{#2}{}}{\ProbOpr{VAR}_{#1}}
{\ifthenelse{\equal{#1}{}}{\ProbOpr{VAR}\left[#2\right]}{\ProbOpr{VAR}_{#1}\left[#2\right]}}} % Expectation: syntax: V{1}{2} = V_1[2], V{}{2}=V[2], V{1}{} = V_1
\DeclareMathOperator*{\argmax}{argmax}

\newcommand{\vm}{\vct{m}}

\newcommand{\vc}{\vct{c}}

\newcommand{\vh}{\vct{h}}

\newcommand{\vvs}{\vct{s}}

\newcommand{\vu}{\vct{u}}
\newcommand{\vv}{\vct{v}}
\newcommand{\vw}{\vct{w}}
\newcommand{\vx}{{\vct{x}}}

\newcommand{\cM}{\cst{M}}
\newcommand{\cN}{\cst{N}}

\newcommand{\sV}{\mathcal{V}}

\newcommand{\vphi}{\vct{\phi}}

\newcommand{\vtheta}{\vct{\theta}}

%% file: imports.tex
\usepackage{xspace}
\usepackage{mathtools}

\usepackage{listings}

\usepackage{epsfig}
\usepackage{graphicx}
\usepackage{float}
\usepackage{wrapfig}
\usepackage{bm}
\usepackage{comment}
\usepackage{verbatim}
\usepackage{multirow}
\usepackage{makecell}
\usepackage{array}
\usepackage{balance}
\usepackage{url}
\usepackage{booktabs}
\usepackage{etoolbox,siunitx}
\usepackage{calc}
\usepackage{pifont,hologo}

\usepackage{arydshln}
\usepackage[utf8]{inputenc} % allow utf-8 input
\usepackage[T1]{fontenc}    % use 8-bit T1 fonts
\usepackage{url}            % simple URL typesetting
\usepackage{amsfonts}       % blackboard math symbols
\usepackage{nicefrac}       % compact symbols for 1/2, etc.
\usepackage{microtype}      % microtypography
\usepackage[american]{babel}
\usepackage{enumitem}
\usepackage{siunitx}

%% file: misc_commands.tex
\usepackage{nicefrac}
\robustify\bfseries

\newcommand{\eat}[1]{{}}

\newcommand\mypara[1]{\vspace{1mm}\noindent\textbf{#1}}

\makeatletter
\DeclareRobustCommand\onedot{\futurelet\@let@token\@onedot}
\def\@onedot{\ifx\@let@token.\else.\null\fi\xspace}
\def\eg{\emph{e.g}\onedot} 
\def\ie{\emph{i.e}\onedot}

\def\etal{\emph{et al}\onedot}
\makeatother
\newcommand{\cmark}{\text{\ding{51}}}
\newcommand{\xmark}{\text{\ding{55}}}

\makeatletter
\@namedef{ver@everyshi.sty}{}
\makeatother
\usepackage[colorinlistoftodos,linecolor=gray,backgroundcolor=none,bordercolor=red,textsize=small,textwidth=16mm]{todonotes}
\definecolor{fscolor}{rgb}{0, 0, 1}
\newcommand{\fsnote}[1]{{}}

%% file: 1_intro.tex
% !TEX root = main_cvpr21.tex
\section{Introduction}
\label{sec:introduction}

With over 70\% of the current internet traffics being video data~\cite{networking2016cisco}, a growing number of videos are being created, shared, and consumed over time. To effectively and efficiently search, browse, and navigate video contents, an intelligent system needs to understand the rich and complex semantic information in them. For this type of use cases, the recently proposed task of \emph{moment localization in video corpus} (MLVC) highlights several challenges in semantic understanding of videos~\cite{escorcia2019temporal,lei2020tvr}. The goal of MLVC is to find a video segment that corresponds to a text query from a corpus of \emph{untrimmed and unsegmented} videos, with a significant amount of variation in factors such as the type of contents, lengths, visual appearance, quality, and so on.

This task can be seen as ``finding a needle in the haystack''. It  is different from searching videos with broad queries such as genres or names of the artists. In contrast, the text query needs to be semantically congruent to a relatively short segment in a much longer target video. For example, the query ``LeBron James shot over Yao Ming'' matches only a few seconds of clips in a game of hours long. Thus, MLVC requires semantic understanding of videos at a more fine-grained level than video retrieval, which typically only targets the whole video. Furthermore, finding the corresponding segment for a text query requires combing through all videos in a corpus and all possible segments in each video. For a large  corpus with long videos, it is not feasible to have such computational complexity that depends on the square of the (averaged) number of frames.

In this paper, we address this challenge by representing videos at multiple scales of granularity.
At the coarse-grained level, the representation captures semantic information in a video over long temporal spans (\emph{e.g.}, clips), allowing us to retrieve the most relevant set of videos for a text query. At the fine-grained level, the representation captures semantic information in short temporal spans (\emph{e.g.}, frames) to allow for precise localization of the most relevant video segments among the retrieved videos.

\begin{figure*}[t]
    \centering
    \small
    \includegraphics[width=\textwidth]{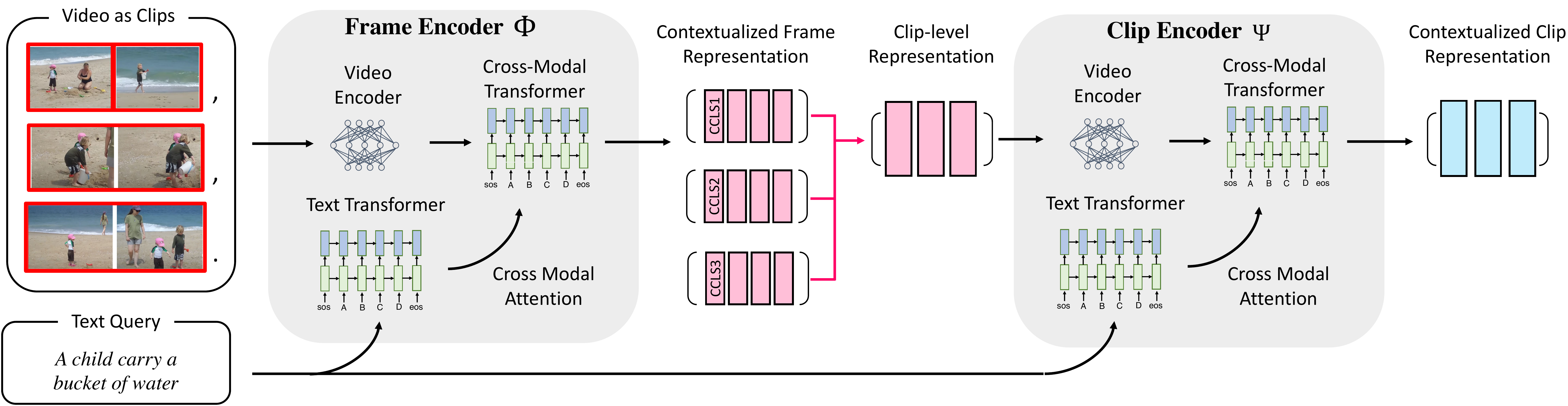}
    \caption{\small Overview of the {\ourmodel} model. The model contains two cross-modal encoders, a frame encoder and a clip encoder on top of it. The outputs of the model are contextualized frame-level and clip-level features, which are used by downstream task-specific modules, \eg video retrieval and temporal localization.}
    \label{fig:model_arch}
\end{figure*}

We propose a novel hierarchical multi-modal encoder ({\ourmodel}) to implement this idea. \ourmodel uses cross-modal attention  to combine the information between the text and visual modalities. The cross-modal learning occurs hierarchically at 3 scales: frame, clip, and video (as a whole). Frames are the most fine-grained building blocks of a video. Each clip consists of a non-overlapping set of frames with equal length, and is in turn the building block of the final video-level representation.
The architecture of the model is illustrated in Fig.~\ref{fig:model_arch}.
The frame-level representation is obtained from a text-visual cross-modal encoder operated on video frames, while the clip-level representation is built upon the frame-level representation with a similar encoder.

The introduction of clip-level representation encoder is important as it allows us to capture both coarse- and fine-grained semantic information. In contrast, existing approaches for MLVC \citep{lei2020tvr,escorcia2019temporal} and other visual-language tasks~\cite{frome2013devise,kiros2014unifying,faghri2017vse++,chen2020learning} typically  pack information of different granularity into a single vector embedding, making it hard to balance the differing demands between retrieving a long video and localizing a short segment.

We apply {\ourmodel} to MLVC task on two large-scale datasets, ActivityNet Captions~\cite{krishna2017dense} and TVR~\cite{lei2020tvr}. We train it with a multi-tasking approach combining three objectives: video retrieval, temporal localization, and an auxiliary masked language modeling.
Our experiments demonstrate the efficacy of {\ourmodel} and establish state-of-the-art performance on all the tasks simultaneously---video retrieval, moment localization in single video and moment localization in video corpus.
To better understand the inner-workings of our model, we compare it with a strong {\ourflat} baseline, a video encoder without any hierarchical representation.
Since the longer videos tend to be less homogeneous, it becomes decidedly important to represent the videos at multiple levels of granularity. Our analysis shows that the performance of a {\ourflat} baseline declines, when the number of frames irrelevant to the text query increases.
On the other hand, the performance of our proposed {\ourmodel} model is robust to the length of the videos, showing that our hierarchical approach is not affected by the increase of irrelevant information and can flexibly handle longer videos.

\mypara{Our contributions} are summarized as follows:
\begin{itemize}[topsep=0.1cm,itemsep=0cm] % leftmargin=*,
    \item We propose a novel model architecture {\ourmodel} that represents videos hierarchically and improves video modeling at long-term temporal scales.
    \item We demonstrate the efficacy of {\ourmodel} on two large-scale datasets, \ie, ActivityNet Captions and TVR, outperforming previous state-of-the-art methods.
    \item We carry out a detailed analysis of {\ourmodel} and show that it particularly improves the performance of video retrieval over strong baselines on long videos.
    \item We conduct a thorough ablation study to understand the effects of different design choices in our model.
\end{itemize}

%% file: 2_related.tex
% !TEX root = main_cvpr21.tex

\section{Related Work}
\label{sec:related}

Most existing MLVC approaches consider text-based video retrieval \cite{xu2015video2txt,dong2016word2visualvec,venugopalan2015vid2text,Pan_2016_CVPR,mithun2018crossmodal} and temporal localization \cite{hendricks2017localizing,gao2017tall,xu2018text,liu2018moment,chen2018temporally,regneri-etal-2013-grounding} as separate tasks.

\mypara{Video Retrieval} (VR) is a task that ranks candidate videos based on their relevance to a descriptive text query. Standard cross-modal retrieval methods~\cite{krishna2017dense,venugopalan2014translating} represent both video and text as two holistic embeddings, which are then used for computing the similarity as the ranking score.
When the text query is a lengthy paragraph, hierarchical modeling is applied to both modalities separately \cite{zhang2018cross,shao2018find}, leading to a significant improvement on the performance of text-based video retrieval.
Different from prior work, in this study we consider a more realistic problem where we use a single query sentence that describes only a small segment to retrieve the entire video.
For instance, the text query ``\emph{Add the onion paste to the mixture}'' may corresponds to a  temporal segment of a few seconds in a long cooking video.

\mypara{Temporal Localization} (TL) aims at localizing a video segment (usually a short fraction of the entire video) described by a query sentence inside a video. Two types of methods have been proposed to tackle this challenge, namely the top-down (or proposal-based) approach \cite{hendricks2017localizing,xu2018text,gao2017tall} and the bottom-up (or proposal-free) approach~\cite{chen2018temporally,lu2019debug,chen2020rethinking,yuan2019semantic,yuan2019find}. The top-down approach first generates multiple clip proposals before matching them with a query sentence to localize the most relevant clip from the proposals. The bottom-up approach first calculates a query-aware video representation as a sequence of features, then predicts the start and end times of the clip described by the query.

\mypara{Moment Localization in Video Corpus} (MLVC) is first proposed by
Escorcia~\etal~\cite{escorcia2019temporal}. They consider a practical setting where they require models to jointly retrieve videos and localize moments corresponding to natural language queries from a large collection of untrimmed and unsegmented videos. They devised a top-down localization approach that compares text embeddings on uniformly partitioned video chunks. Recently, Lei~\etal~\cite{lei2020tvr} proposed a new dataset, TVR, that considers a similar task called Moment Retrieval in Video-Subtitle Corpus, which requires a model to align a query text temporally with a video clip, using multi-modal information from video and \emph{subtitles} (derived from Automatic Speech Recognition or ASR).

%% file: 3_method.tex
% !TEX root = main_cvpr21.tex
\section{Method}
\label{sec:method}

We first describe the problem setting of MLVC and introducing the notations in \S\ref{subsec:problem}. In \S\ref{subsec:decompose}, we describe a general strategy of decomposing MLVC into two sub-tasks, VR and TL~\cite{xu2019multilevel,lu2019debug}. The main purpose  is to reduce computation and to avoid the need to search all possible segments of all videos. In \S\ref{subsec:hammer}, we present a novel HierArchical Multi-Modal EncodeR ({\ourmodel}) model and describe how it is trained in \S\ref{subsec:learning}. Finally, we describe key details for inference in \S\ref{subsec:inference}.

\subsection{Problem Setting and Notations}
\label{subsec:problem}

We represent a video $\vv$ as a sequence of $\cN$ frames $\{\vx_t | t = 1, \ldots, \cN \}$, where $\vx_t$ is a visual feature vector representing the $t$-th frame. Given a text query $\vh$ (\eg, a sentence),  our  goal is to learn a parameterized function (\ie, neural networks) that accurately estimates the conditional probability $p(\vvs | \vh)$, where $\vvs$ is a video segment
given by $\vvs = \{\vx_t | t = t^{\cst{s}}, \ldots, t^{\cst{e}} \}$. $t^{\cst{s}}$ and $t^{\cst{e}}$ stand for the indices of the starting and the ending frames of the segment in a video $\vv$. Note that for a video corpus $\sV$ with an average length of $\cN$ frames, the number of all possible segments is $O(|\sV | \times \cN^2)$. Thus, in a large corpus, exhaustive search for the best segment $\vvs$ corresponding to $\vh$ is not feasible. In what follows, we describe how to address this challenge.

To localize the moment $\vvs$ that best corresponds to a text query $\vh$, we need to identify
\begin{align}
    \vvs^* = \argmax_{\vvs} p( \vvs|\vh)
           = \argmax_{\vvs} \sum_{\vv} p(\vvs|\vv, \vh) p(\vv|\vh)
\end{align}
Note that the conditional probability is factorized into two components. If we assume $\vvs$  uniquely belongs to only one video in the corpus $\sV$, then the marginalization over the video $\vv$ is vacuous and can be discarded. This leads to
\begin{equation}
    \max_{\vvs} p(\vvs|\vh) = \max_{\vv} \max_{\vvs \in \vv} p(\vvs|\vv, \vh) p(\vv|\vh).
    \label{eFactor}
\end{equation}
The training data are available in the form of  $(\vh^{(i)}, \vv^{(i)}, \vvs^{(i)})$ where $\vvs^{(i)} \subset \vv^{(i)}$ is the matched segment to the query $\vh^{(i)}$.

\subsection{Two-Stage MLVC: Retrieval and Localization}
\label{subsec:decompose}

As aforementioned, this inference of Eq.~\eqref{eFactor} is infeasible for large-scale corpora and/or long videos. Thus, we approximate it by
\begin{equation}
    \vvs^* \approx \argmax_{\vvs \in \vv^*} p(\vvs|\vv^*,\vh), \text{ with } \vv^* = \argmax_{\vv} p(\vv|\vh)
    \label{eApprox}
\end{equation}
This approximation allows us to build two different learning components and stage them together to solve MLVC. This approach has been applied in a recent work on the task~\cite{lei2020tvr}. We give a formal summary below.

\mypara{Video Retrieval (VR)} identifies the best video $\vv^*$ by minimizing the negative log-likelihood of $p(\vv|\vh)$
\begin{equation}
    \ell^{\textsc{vr}} = -\sum_i \log p( \vv^{(i)}|\vh^{(i)})
\end{equation}
where $\vv^{(i)}$ is the ground-truth video for the text query $\vh^{(i)}$. This is a rather standard (cross-modal) retrieval problem, which has been widely studied in the literature. (See \S\ref{sec:related} for some references.)

\mypara{Temporal Localization (TL)} models $p(\vvs|\vv, \vh)$. While it is possible to model $O(\cN^2)$ possible segments in a video with $\cN$ frames, we choose to model it with the probabilities of identifying the correct starting ($t^{\cst{s}}$) and ending ($t^{\cst{e}}$) frames:
\begin{align}
    p(\vvs | \vv, \vh) &\approx p(t^{\cst{s}} | \vv, \vh) \cdot p(t^{\cst{e}} | \vv, \vh) \cdot \mathbb{I}[t^{\cst{e}} > t^{\cst{s}}] \label{eqn:tl}
\end{align}
Here, we consider $t^{\cst{s}}$ and $t^{\cst{e}}$ to be independent to efficiently approximate $p(\vvs | \vv, \vh)$. The indicator function $\mathbb{I}[\cdot]$ simply stipulates that the ending frame needs to be after the starting frame.

To model each of the factors, we treat it as a frame classification problem, annotating each frame with one of the three possible labels: \texttt{\underline{B}EGIN} and \texttt{\underline{E}ND} marks the starting and ending frames respectively, with all other frames as  \texttt{\underline{O}THER}. We denote this as \texttt{B}, \texttt{E}, \texttt{O} classification scheme. During  training, we optimize (the sum of) the frame-wise cross-entropy between the model's predictions and the labels. We denote the training loss as
\begin{equation}
        \ell^{\textsc{TL}} = -\sum_i \sum_t f_t^{(i)}\log p(y_t^{(i)}|\vv^{(i)}, \vh^{(i)}),
\end{equation}
where $f_t^{(i)}$ is the true label for the frame $\vx_t$ of the video $\vv^{(i)}$, and $y_t^{(i)}$ is the corresponding prediction of the model.

This type of labeling schemes have been widely used in the NLP community, for example, recently for span-based question and answering~\cite{joshi2020spanbert,fevry2020entities}.

\subsection{HierArchical Multi-Modal Encoder (\ourmodel)}
\label{subsec:hammer}

Our first contribution is to introduce the hierarchical modeling approach to parameterize the conditional probability $p(\vv|\vh)$ for the VR sub-task and the labeling model $p(y|\vv, \vh)$ for the TL sub-task. In the next section, we describe novel learning algorithms for training our model.

\mypara{Main idea}  Video and text are complex and structural objects. They are naturally in ``temporally'' linear orders of frames and words. More importantly, semantic relatedness manifests in both short-range and long-range contextual dependencies. To this end, \ourmodel  infuses  textual and visual information hierarchically at different temporal scales.  Figure~\ref{fig:model_arch} illustrates the architecture of \ourmodel. A key element here is to introduce cross-modal attention at both the frame level and the clip level.

\mypara{Clip-level Representation}  We introduce an intermediate-level temporal unit with a fixed length of $\cM$ frames, and refer to them as a clip $\vc_k = \{\vx_t | t = (k-1)\cdot \cM,\ldots, k\cdot \cM - 1\}$, where $k=1,\ldots, \lceil \cN / \cM \rceil $. As such, a video can also be hierarchically organized as a sequence of non-overlapping video clips $\vv = \{ \vc_k | k = 1,\ldots, \lceil \cN / \cM \rceil\}$.  $\cM$ is a hyper-parameter to be adjusted on different tasks and datasets. We emphasize while sometimes segments and clips are used interchangeably, we refer to ``segment'' as a set of frames that are also the visual grounding of a text query, and ``clip'' as a collection of temporally contiguous frames. We treat them as holding memory slots for aggregated lower-level semantic information in frames.

\mypara{Cross-modal Transformers} \ourmodel has two cross-modal Transformers. At the frame-level, the frame encoder $\Phi$  takes as input both the frame sequence of a video clip and the text sequence of a query, and outputs the contextualized visual frame features $\{\Phi(\vx_t; \vc_k, \vh) \}$ for each clip $\vc_k$. The frame encoder $\Phi$ encodes the local and short-range contextual dependencies among the frames of the same clip.

We also introduce a \emph{Clip CLS Token} ($\texttt{CCLS}_k$) for each $\vc_k$~\cite{lu2019vilbert}. The contextual embedding of this token gives the representation of the clip:
\begin{equation}
  \vphi_k = \Phi(\texttt{CCLS}_k; \vc_k, \vh)
\end{equation}
Contextual embeddings for all clips are then fed into a higher-level clip encoder $\Psi$, also with cross-modal attention to the input text, yielding a set of contextualized clip representation
\begin{equation}
  \Psi_{\vv} = \{\Psi(\phi_k; \vv, \vh) \mid k = 1,\ldots,\lceil \cN / \cM \rceil\}.
\end{equation}
Note that $\Psi_{\vv}$ now encodes the global and longer-range contextual dependencies among all frames (through clips)\footnote{Alternatively, we can summarize it (into a vector, in lieu of the set) through various reduction operations such as pooling or introducing a video-level \emph{CLS token} $\texttt{VCLS}$.}.

To summarize, our model has 3 levels of representations: the contextualized frames $\{\Phi(\vx_t; \vc_k, \vh)\}$, the clips $\{\vphi_k\}$, and the entire video $\Psi_{\vv}$.  Next, we describe how to use them to form our learning algorithms.

\subsection{Learning \ourmodel for MLVC}
\label{subsec:learning}

The different levels of representation allows for the flexibility for modeling the two subtasks (VR and TL) with semantic information across different temporal scales.

\mypara{Modeling Video Retrieval} We use the contextualized clips to compute the video-query compatibility score for a query $\vh$ and its corresponding video $\vv$. In order to retrieve the likely relevant videos as much as possible, we need a coarse-grained matching that focuses more on higher-level semantic information.

Specifically, we identify the best matching among all clip embeddings $\Psi(\vc_k ; \vv, \vh)$ and use it as the matching score for the whole video:
\begin{align}
    p(\vv | \vh) \; {\propto} \; f(\vv, \vh) = \max_k\left( \{ \vtheta_{\textsc{vr}}^\top \cdot \Psi(\vphi_k ; \vv, \vh) \} \right) \label{eqn:vr_head}
\end{align}
where $\vtheta_{\textsc{vr}}$ is a linear projection to extract the matching scores\footnote{An alternative design is to pool all $\Psi(\vphi_k ; \vv, \vh)$ and then perform a linear projection. However, this type of polling has a disadvantage that a short but relevant segment -- say within a clip -- can be overwhelmed by all other clips. Empirically, we also find the current formalism works better. A similar finding is also discovered in~\cite{zhang2018cross}.}. The conditional probability is normalized with respect to all videos in the corpus (though in practice, a set of positive and negative ones).

\mypara{Modeling Temporal Localization} As in the previous section, we treat localization as classifying a frame into \texttt{B}, \texttt{E}, or \texttt{O}:
\begin{equation}
    p( y_t | \vv, \vh) \approx  p(\vc_k | \vv, \vh) \cdot p( y_t | \vc_k, \vh)
\end{equation}
Note that each frame can belong to only one clip $\vc_k$ so there is no need to marginalize over $\vc_k$.  The probability $p(\vc_k|\vv, \vh)$ measures the likelihood of $\vc_k$ containing a label $y_t$ in one of its frames. The second factor measures the likelihood that the specific frame $\vx_k$ is labeled as $y_t$. Clearly, these two factors are on different semantic scales  and are thus modeled separately:
\begin{align}
    p(\vc_k|\vv, \vh) & \propto \vu^\top \cdot\; [\Psi(\phi_k; \vv, \vh),\; \Psi(\texttt{TCLS}; \vv, \vh)\;]\\
    p(y_t|\vc_k, \vh) & \propto \vw_{y_t}^\top  \cdot [\Phi(\vx_t; \vc_{k}, \vh), \Phi(\texttt{TCLS}; \vc_k, \vh)]
\end{align}
where \texttt{TCLS} is a text \texttt{CLS} token summarizing the query embedding.

\mypara{Masked Multi-Modal Model} Masked language modeling has been  widely adopted as a pre-training task for language modeling~\cite{lu2019vilbert,sun2019videobert,devlin2018bert}. The main idea is to backfill a masked text token from its contexts, \ie, the other tokens in a sentence.

The multi-modal modeling task in this paper can similarly benefit from this idea. During training, we mask randomly some text tokens. We expect the model to achieve two things: (1) using the partially masked text query to retrieve and localize which acts as a regularization mechanism; (2) better text grounding by recovering the masked tokens with the assistance of the multimodal context, i.e., both the textual context and the visual information in the frames and the clips.

To incorporate a masked query to the loss functions $\ell^{\textsc{vr}}$ and $\ell^{\textsc{tl}}$ of the model we apply $\vh \otimes (\mathbf{1}-\vm)$ to replace $\vh$, where $\vm$ is a binary mask vector for text tokens, $\mathbf{1}$ is a one-valued vector of the same size, and $\otimes$ indicates element-wise multiplication. We introduce another loss to backfill the missing tokens represented by $\vh \otimes \vm$:
\begin{equation}
    \ell^{\textsc{mask}} = - \log p( \vh \otimes \vm | \vv, \vh \otimes (\mathbf{1}-\vm))
\end{equation}
This probability is computed using both $\Phi(\cdot)$ for  frames and  $\Psi(\cdot)$ for  clips.

\mypara{Multi-Task Learning Objective} We use a weighted combination of video retrieval, moment localization, and masked multi-modal modeling objectives as our final training objective:
\begin{align}
  \ell = \mathbb{E}_{\vm}\left[\lambda^{\textsc{vr}} \cdot \ell^{\textsc{vr}} + \lambda^{\textsc{tl}} \cdot \ell^{\textsc{tl}} + \lambda^{\textsc{mask}}\cdot \ell^{\textsc{mask}} \right],
\end{align}
where the expectation is taken with respect to random masking.
Since the VR and TL subtasks share the same model and output representations, the final objective needs to balance different goals and is multi-tasking in nature.
We provide a detailed ablation study in \S\ref{sec:experiments} to analyze the choice of weights.

\subsection{Two-stage Inference with {\ourmodel}}
\label{subsec:inference}

For the model inference of {\ourmodel}, we perform two sequential stages, \ie, video retrieval and temporal localization, to accomplish the task of moment localization in video corpus. For video retrieval, we use {\ourmodel} and the linear regressor to compute pairwise compatibility scores as in Eq.~\eqref{eqn:vr_head} with respect to the text query $\vh$ and all videos $\sV$ in the corpus. Next, we perform temporal localization on the top ranked videos. Specifically, we predict the start and end frame with  {\ourmodel} to localize the temporal segment $\vvs$ following  Eq.~\eqref{eqn:tl}. Then we greedily label the frame with the maximum $p(t^{\cst{s}} | \vv, \vh)$ as the start frame and maximum $p(t^{\cst{e}} | \vv, \vh)$ as the end frame. Here we have an additional constraint to consider --- the predicted end frame must appear after the start frame prediction.
This two-stage inference reduces the complexity to $O(|\sV| + \cN)$, which is significantly better comparing to the $O(|\sV| \cdot \cN^2)$ complexity of ~\cite{escorcia2019temporal}.

%% file: 4_exp.tex
% !TEX root = main_cvpr21.tex

\section{Experiments}
\label{sec:experiments}

In this section, we perform experiments with the proposed \ourmodel model. We first introduce the datasets and setups of our experiments in \S\ref{subsec:exp:setup}. Next, we present the main results of the {\ourmodel} model in \S\ref{subsec:exp:mainresult}, contrasting against a strong baseline \ourflat as well as other existing methods. We then confuct a thorough ablation study in \S\ref{subsec:exp:ablation} to evaluate the importance of various design choices for the {\ourmodel} model. Finally, we carry out qualitative analysis of our model to better demonstrate its behaviour.

\subsection{Experimental Setups}
\label{subsec:exp:setup}

\mypara{Datasets} We experiment on two popular MLVC datasets:
\begin{itemize}[leftmargin=*,topsep=0.1cm,noitemsep]
    \item \textbf{ActivityNet Captions}~\cite{krishna2017dense} contains $\sim$20K videos, each has 3.65 temporally localized query sentences on average. The mean video duration is 180 seconds and the average query length is 13.48 words, which spans over 36 seconds of the video. There are 10,009 videos for training and 4,917 videos for validation (val\_1 split). We follow prior work~\cite{escorcia2019temporal,hendricks2018localizing} to train our models and evaluate them on the val\_1 split.
    \item \textbf{TVR}~\cite{lei2020tvr} contains $\sim$22K videos in total, of which $\sim$17.5K videos are in the training set and 2,180 are in the validation set. The dataset contains videos from movies, cartoons, and TV-shows. The videos are on average 76.2 seconds long and contain 5 temporally localized sentences per video. The moments in the videos are 9.1 seconds long and described by sentences containing 13.4 words on average. We make use of the subtitle (ASR) features together with the video feature in TVR dataset, following prior works~\cite{lei2020tvr,li2020hero}.
\end{itemize}

We make use of multiple popular choices of video features on these two datasets as existing literature~\cite{escorcia2019temporal,hendricks2018localizing,lei2020tvr}, which includes the appearance-only features (ResNet152~\cite{he2016deep} pre-trained on ImageNet~\cite{deng2009imagenet}), spatio-temporal features (I3D~\cite{carreira2017quo} pre-trained on Kinetics~\cite{kay2017kinetics}), and their combinations. We present the details of feature preparation in \SM.

\mypara{Evaluation Metrics}
We use different evaluation metrics for different video understanding tasks:
\begin{itemize}[leftmargin=*,topsep=0cm,noitemsep]
    \item \textbf{Video Retrieval} ({VR}) We report Recall@$k$ and Median Rank (MedR or MedRank) as the evaluation metrics for {Video Retrieval} as suggested in the literature.
    \item \textbf{{Temporal Localization}} (TL) We report both mean IoU (mIoU) and average precision with IoU=\{0.3, 0.5, 0.7\} as the evaluation metrics. Here, IoU measures the Intersection over Union between the ground truth and predicted video segments, \ie,  the localization accuracy.
    \item \textbf{Moment Localization in Video Corpus} (MLVC) We use Recall@$k$ with IoU=$p$ for the main evaluation metrics~\cite{escorcia2019temporal,lei2020tvr}. Specifically, we measure whether the correct localized segment exists in the top $k$ of the ranked videos. Here, a localized segment is correct if it overlaps with the ground truth segment {over} an IoU of \{0.5, 0.7\}.
\end{itemize}

\mypara{Baseline and the {\ourmodel} Models} In {\ourmodel}, we use two encoders, \ie, the frame and clip encoders, with multiple Transformer~\cite{Vaswani2017Attention} layers to represent the visual (and ASR) features as well as the text query features (details in Figure~\ref{fig:model_arch}). Each encoder contains 1 layer of Transformer for visual input, 5 layers of Transformers for the text query input, and 1 layer of cross-modal Transformer between the visual and text query inputs. When ASR is provided (\ie, in TVR), we add one additional Transformer layer to incorporate the ASR input, with another cross-modal Transformer layer that cross-attends between the query input and ASR features. The processed ASR and visual features are concatenated. Meanwhile, we design a {\ourflat} model as a strong baseline. The {\ourflat} model has a similar architecture as {\ourmodel}, except that it only uses the frame encoder to capture the visual (and ASR) features.  We provide complete details about the architectural configurations and model optimization in the \SM.

\input{tables/main_mlvc_results}

\footnotetext[1]{We compare against their model {without} large-scale pre-training for fair comparison.}

\subsection{MLVC Experiments}
\label{subsec:exp:mainresult}

\mypara{Main Results} Table~\ref{tab:main} presents a comparison between the proposed {\ourmodel} and other methods on the two MLVC benchmarks. We observe that, irrespective of the feature types, \ourmodel outperforms \ourflat noticeably, which in turn outperforms most published results on both datasets. On ActivityNet, we observe that models using I3D features (denoted as I) outperform their counterparts with ResNet152 (denoted as R) features, by a significant margin. It indicates the importance of spatio-temporal feature representation in the MLVC tasks.

Meanwhile, we note that our {\ourflat} model outperforms the baselines on the TVR dataset, which is mainly due to the introduction of the cross-modal Transformer between query and visual$+$ASR features (see \S\ref{subsec:exp:ablation} for a detailed study). On both datasets, {\ourmodel} establishes the new state-of-the-art results for the MLVC task (without using additional data).
This result shows a clear benefit of hierarchical structure modeling in video for the MLVC task.

\input{tables/main_vr_results}
\input{tables/main_tl_results}

Table~\ref{tab:hier_vr_only} and~\ref{tab:hier_tl_only} contrast \ourmodel to the \ourflat model in more details by comparing their performance on the tasks of video retrieval and temporal localization separately. The results are reported on the ActivityNet with models using the I3D features. In both cases, \ourmodel achieves significantly better performance than the baseline \ourflat model.

\mypara{Comparing Models on Videos of Different Duration} We discuss the potential reasons for {\ourmodel} to outperform the {\ourflat} model.
Since {\ourmodel} learns video representation at multiple granularities, we hypothesize that it should be able to focus on the task-relevant parts of a video without getting distracted by irrelevant parts.
Specifically for the task of sentence-based video retrieval which requires matching the relevant frames in the video with the text query, {\ourmodel} would be less sensitive to the presence of non-matching frames and hence be robust to the length of the video. To verify this, we analyze {\ourmodel}'s performance on videos with different lengths for the task of video retrieval and temporal localization.

\begin{figure}[t]
    \centering
    \small
    \includegraphics[width=0.475\textwidth]{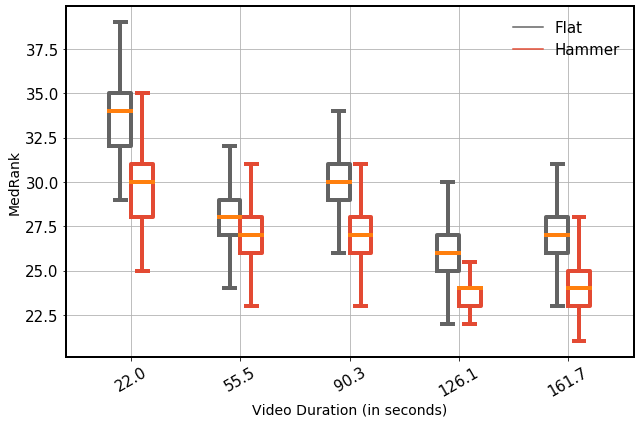}
    \vspace{-15pt}
    \caption{\small Comparison of {{Video Retrieval}} performances under different video duration. Results are reported in Median Rank (MedRank) on the ActivityNet Captions (\textbf{Lower is better}).}
    \label{fig:ablation_vr}
\end{figure}

\begin{figure}[t]
    \centering
    \small
    \includegraphics[width=0.475\textwidth]{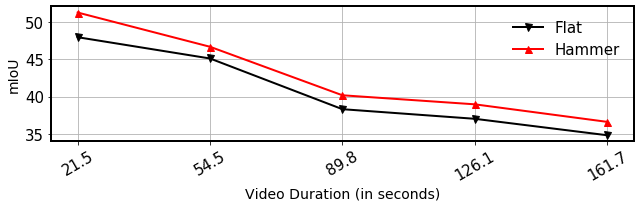}
    \vspace{-15pt}
    \caption{\small Comparison of {{Temporal Localization}} performances under different video duration. Results are reported in Mean IoU (mIoU) on the ActivityNet Captions (\textbf{Higher is better}).}
    \label{fig:ablation_tl}
\end{figure}

We compare the performance of {\ourmodel} and {\ourflat} on videos with different durations for the video retrieval task in Fig.~\ref{fig:ablation_vr}. The metric used for comparison is the median rank where lower numbers indicate better performance. Firstly, it can be observed that while the performance of {\ourflat} model is inconsistent (e.g., performance on longest videos is worse than second-to-longest videos), the {\ourmodel} model's performance consistently improves with the length of the video. Secondly, the performance of the {\ourmodel} model is best for the longest videos in the dataset. Finally, while both the models perform sub-optimally on the shortest videos, {\ourmodel} still outperforms {\ourflat} for those videos.

We further compare the temporal localization performance of {\ourmodel} and {\ourflat} models in Fig.~\ref{fig:ablation_tl}. The results are reported using mean IoU, where higher numbers indicate better performance. It shows that {\ourmodel} constantly achieves higher performance than {\ourflat} across all videos irrespective of their length.

Overall, the analysis shows clear advantage of using {\ourmodel} over {\ourflat} which is especially profound for longer videos, hence supporting our central modeling argument.

\subsection{Ablation Studies and Analyses}
\label{subsec:exp:ablation}

In this section, we evaluate the effectiveness of learning objectives and various design choices for  {\ourmodel}. We note that all the ablation studies in this section are conducted on ActivityNet Captions using the I3D features.

\subsubsection{Learning Objectives}
\fsnote{This section's text needs to be compressed}

\input{tables/ablation_objectives}

As aforementioned, {\ourmodel} is optimized with three objectives jointly namely video retrieval (VR), temporal localization (TL), and masked language modeling (MLM). We study the contribution of different objectives, reported in Table~\ref{tab:ablation_objectives}. It is worth to note that here we differentiate the MLM objective applied to the frame encoder (denote as \texttt{FM}) and the clip encoder (denote as \texttt{CM}).
Firstly, the objectives of VR and TL are complementary to each other and jointly optimizing the two surpasses the single task performance on both the tasks simultaneously. Secondly, \texttt{CM} and \texttt{FM} applied individually benefits both VR and TL tasks with their usage in unison resulting in best performance. This verifies the effectiveness of MLM objective to improve the text representation. Finally, the best performance is achieved by combining all the objectives, hence proving the complimentary nature of all of them.

\input{tables/ablation_loss_weights}

\mypara{Weights of Different Objectives} We also conduct detailed experiments to investigate the influence of different objectives' weights ($\lambda^{\textsc{vr}}$, $\lambda^{\textsc{tl}}$, and $\lambda^{\textsc{mlm}}$). Table~\ref{tab:ablation_loss_weights} shows that it is important to balance the weights between VR and TL. The best performance is achieved when the weight for VR and TL is set to $1:5$. For MLM, we find that the best loss weight is 0.1, and thus use this value through all our experiments.

\subsubsection{Evaluate Design Choices of the {\ourmodel}}
We study the importance of a few design choices in the {\ourmodel} model. Specifically, we evaluate the following:
\begin{itemize}[leftmargin=*,noitemsep,topsep=0.1cm]
    \item Effect of the \emph{cross-modal Transformer} layer
    \item Effect of different \emph{clip lengths} for the clip representation
    \item Effect of \emph{parameter sharing} for frame and clip encoders
    \item Effect of an additional clip-level \emph{position embedding}
\end{itemize}
We present the results and discussion on these experiments in the following paragraphs.

\input{tables/ablation_vse}

\begin{figure*}[tbh]
    \centering
    \small
    \includegraphics[width=\textwidth]{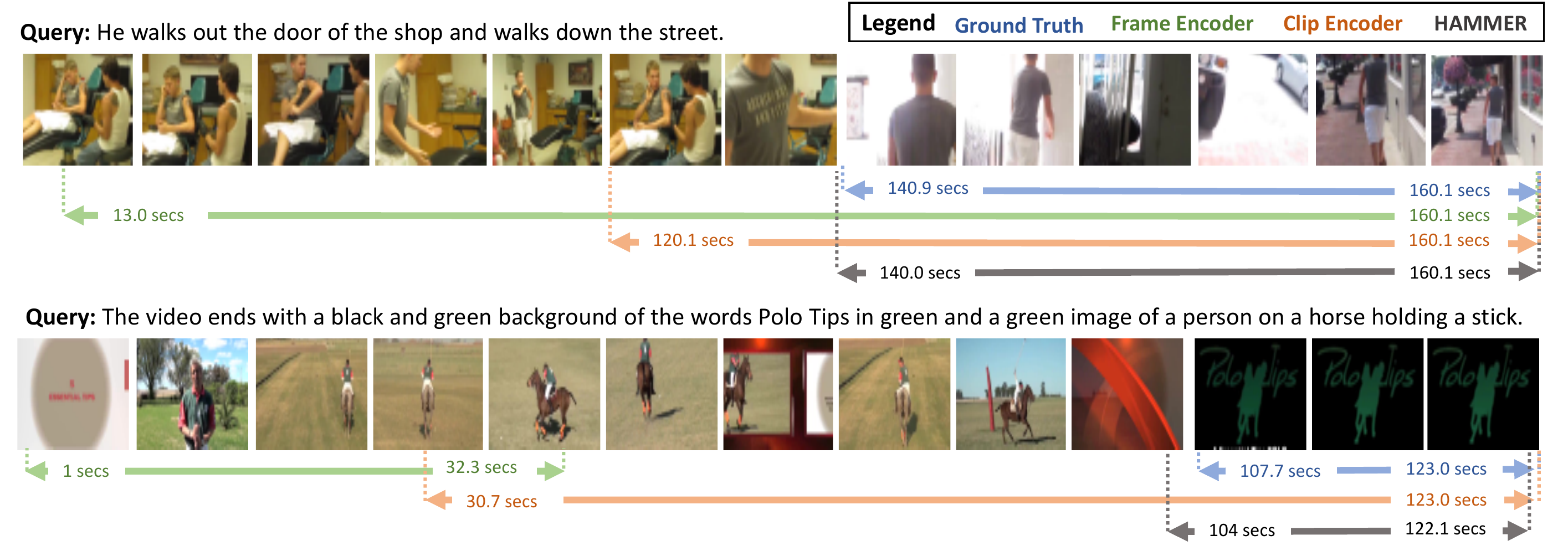}
    \vspace{-10pt}
    \caption{Illustration of temporal localization using different hierarchies of {\ourmodel} as well as the final {\ourmodel} model}
    \label{fig:qualitative_result}
\end{figure*}

\mypara{Cross-Modal Transformer is Essential.} Both frame and clip encoders contain one layer of cross-modal (X-modal) Transformer between text query and video inputs. To verify its effectiveness, we compare with an ablation model without this layer. Table~\ref{tab:ablation_vse} shows almost 100\% relative improvement in the R$1$, R$10$ metrics when using the X-modal Transformer, proving it is essential to the success of {\ourmodel}.

\input{tables/ablation_clip_len}
\mypara{Optimal Length of the Clip-Level Representation.} In {\ourmodel}, recall that a frame encoder takes a clip of fixed length of frames and outputs a clip-level representation. Here, we examine the performance under different lengths of clips, summarized in Table~\ref{tab:ablation_clip_len}. Overall, we observe that the model's performance is robust to the clip length chosen for the experiments, and 32 is the optimal length for the clip representation (with max video length of 128).

\input{tables/ablation_weightshare}

\mypara{Parameter Sharing between Frame/Clip Encoders.}
We also consider whether the frame encoder and the clip encoder in the {\ourmodel} model may share the same set of parameters, as weight sharing could regularize the model capacity and therefore improve the generalization performance. Table~\ref{tab:ablation_weightshare} indicates that, however, untying the encoder weights achieves slightly better performance, potentially thanks to its greater flexibility.

\input{tables/ablation_clippos}

\mypara{Position Embedding for Clip Encoder.} Position embedding is an important model input as it indicates the temporal boundary of each video frame segment. In the {\ourmodel} model, since we also have a clip encoder that takes the aggregated ``Clip CLS'' token as input, it is natural to ask if we need a position encoding for each clip representation. Thus, we compare two models, with and without additional clip position encoding. Table~\ref{tab:ablation_clippos} shows that clip position embedding is indeed important to achieve superior performance.

\subsubsection{Qualitative Visualization}

To better understand the behavior of the {\ourmodel} model, we demonstrate a couple of examples of temporal localization. Figure~\ref{fig:qualitative_result} lists predicted spans from the frame and clip encoder as well as from the entire {\ourmodel}.
%prediction, as well as the predictions from either frame or clip encoders
In both examples, we observe that the frame encoder of {\ourmodel} makes an incorrect prediction of the temporal timestamps, but then corrected by the prediction from the clip encoder. Overall, {\ourmodel} makes more accurate predictions with respect to the ground-truth video segment.

%% file: tables/main_mlvc_results.tex
\begin{table}[t]
    \small
    \tabcolsep 3.5pt
    \centering
    \caption{
        MLVC Results on ActivityNet and TVR datasets
    }
    \vspace{0.1cm}
    \label{tab:main}
    \begin{tabular}{l@{\;}ll@{\;} rrr @{\quad} rrr}
    \toprule
        &   &  & \multicolumn{3}{c}{IoU=0.5} & \multicolumn{3}{c}{IoU=0.7} \\
        \cmidrule(lr){4-6} \cmidrule(lr){7-9}
        & \multicolumn{2}{c}{Model \& Feature} & R1 & R10 & R100 & R1 & R10 & R100 \\
        \midrule
        \multirow{5}{*}[-0.25em]{\rotatebox[origin=c]{90}{ActivityNet}}
        & \textsc{mcn}~\cite{hendricks2017localizing}  & R &  0.02   &   0.18   &   1.26   & 0.01  & 0.09   &  0.70   \\
        & \textsc{cal}~\cite{escorcia2019temporal}  & R & 0.21 & 1.32 & 6.82 & 0.12 &   0.89 & 4.79 \\
        & {\ourflat}  & R & 0.34  & 2.28 & 10.09 & 0.21  &  1.28  & 5.69 \\
        & {\ourmodel} & R & 0.51  & 3.29 & 12.01 & 0.30  &  1.87  & 6.94 \\
        & {\ourflat}  & I & 2.57   & 13.07    & 30.66    & 1.51  & 7.69   & 17.67 \\
        & {\ourmodel} & I & \textbf{2.94}  & \textbf{14.49}  & \textbf{32.49}  & \textbf{1.74} & \textbf{8.75} & \textbf{19.08} \\
        \midrule
        \multirow{5}{*}[0.5em]{\rotatebox[origin=c]{90}{TVR}}
        & \textsc{xml}~\cite{lei2020tvr} & I+R & -- & -- & -- & 2.62 & 6.39 & \textbf{22.00}    \\
        & \textsc{hero}\footnotemark[1]~\cite{li2020hero} & I+R & -- & -- & -- & 2.98 & 10.65 & 18.25     \\
        & {\ourflat}  & I+R & 8.45  & 21.14 & 30.75  & 4.61   & 11.29 & 16.24 \\
        & {\ourmodel} & I+R & \textbf{9.19}  & \textbf{21.28}  & \textbf{31.25}  & \textbf{5.13}  & \textbf{11.38}  & 16.71 \\
    \bottomrule
    \end{tabular}
\end{table}

%% file: tables/main_vr_results.tex
\begin{table}[t]
    \small
    \centering
    \caption{VR results on ActivityNet Captions.}
    \label{tab:hier_vr_only}
    \centering
    \begin{tabular}{lcccc}
    \toprule
    Model          &  R1   & R10   & R100  & MedR$\downarrow$ \\
    \midrule
    {\ourflat}     &  5.37  & 29.14  &  71.64  & 29  \\
    {\ourmodel}    &  \textbf{5.89}  & \textbf{30.98}  &  \textbf{73.38}  & \textbf{26} \\
    \bottomrule
    \end{tabular}
\end{table}

%% file: tables/main_tl_results.tex
\begin{table}[t]
    \small
    \centering
    \caption{TL results on ActivityNet Captions.}
    \label{tab:hier_tl_only}
    \vspace{0.1cm}
    \centering
    \begin{tabular}{lcccc}
    \toprule
    Model & IoU=0.3 & IoU=0.5 & IoU=0.7 & mIoU$\uparrow$ \\
    \midrule
    {\ourflat}     & 57.58 & 39.60 & 22.59 & 40.98 \\
    {\ourmodel}    & \bf 59.18 & \bf 41.45 & \bf 24.27 & \bf 42.68 \\
    \bottomrule
    \end{tabular}
\end{table}

%% file: tables/ablation_objectives.tex
\begin{table}[t!]
    \centering
    \small
    \caption{Ablation study on sub-tasks (\texttt{VR}=Video Retrieval, \texttt{TL}=Temporal Localization, \texttt{FM}=Frame MLM, \texttt{CM}=Clip MLM)}

    \label{tab:ablation_objectives}
    \vspace{0.1cm}
    \centering
    \begin{tabular}{c@{\;}c@{\;}c@{\;}cc@{\;\;\;}c@{\;\;\;}cc@{\;}c@{\;}c}
    \toprule
    \multicolumn{4}{c}{Task} & \multicolumn{3}{c}{Video Retrieval} & \multicolumn{3}{c}{Temporal Localization} \\
    \cmidrule(lr){5-7} \cmidrule(lr){8-10}
    \texttt{VR} & \texttt{TL}& \texttt{FM} & \texttt{CM} & R1 & R10 & R100 & IoU=0.5 & IoU=0.7 & mIoU \\ \midrule
    \cmark & & & & 4.93 & 29.02 & 72.15 & -- & -- & -- \\
    \cmark & & \cmark & & 5.52 & 30.53 & 73.02 & -- & -- & -- \\
    \cmark & & & \cmark & 5.45 & 30.45 & 73.24 & -- & -- & --  \\
    \cmark & & \cmark & \cmark & 5.67 & 30.20 & 72.67 & -- & -- & --\\
    \midrule
    & \cmark & & & -- & -- & -- & 39.02 & 22.74 & 40.28 \\
    & \cmark & \cmark & & -- & -- & -- & 39.27 & 22.04 & 40.30 \\
    & \cmark & & \cmark & -- & -- & -- & 39.13 & 22.38 & 40.51 \\
    & \cmark & \cmark & \cmark & -- & -- & -- & 39.16 & 22.82 & 40.64 \\
    \midrule
    \cmark & \cmark & & & 5.22 & 30.22 & 72.70 & 40.59 & 23.70 & 42.01 \\
    \cmark & \cmark & \cmark & & 5.57 & 30.97 & 73.09 & 41.17 & 24.04 & 42.45  \\
    \cmark & \cmark & & \cmark & 5.85 & 30.82 & \bf 73.54 & 41.30 & 23.94 & 42.43 \\
    \cmark & \cmark & \cmark & \cmark & \bf 5.89 & \bf 30.98 & 73.38 & \bf 41.45 & \bf 24.27 & \bf 42.68 \\
    \bottomrule
    \end{tabular}
\end{table}

%% file: tables/ablation_loss_weights.tex
\begin{table}[t]
    \small
    \tabcolsep 5pt
    \centering
    \caption{Ablation study on task weights (\texttt{VR}=Video Retrieval, \texttt{TL}=Temporal Localization, \texttt{MLM}=Masked Language Model)}
    \vspace{0.1cm}
    \label{tab:ablation_loss_weights}
    \begin{tabular}{@{\;}ccc rrr @{\quad} rrr}
    \toprule
         & & & \multicolumn{3}{c}{IoU=0.5} & \multicolumn{3}{c}{IoU=0.7} \\
        \cmidrule(lr){4-6} \cmidrule(lr){7-9}
$\lambda^{\textsc{vr}}$ & $\lambda^{\textsc{tl}}$ & $\lambda^{\textsc{mlm}}$ & R1 & R10 & R100 & R1 & R10 & R100 \\
        \midrule
 1.0 & 0.1 & 0.1 & 1.65 & 9.18 & 20.81 & 0.87 & 4.88 & 10.50 \\
1.0 & 0.5 & 0.1 & 2.15 & 10.75 & 23.41 & 1.10 & 5.68 & 12.16\\
1.0 & 1.0 & 0.1 & 2.02 & 10.95 & 24.74 & 1.10 & 6.07 &  13.12 \\
1.0 & 5.0 & 0.1 & \textbf{2.94}  & \textbf{14.49}  & \textbf{32.49}  & \textbf{1.74} & \textbf{8.75} & \textbf{19.08} \\
1.0 & 10.0 & 0.1 & 2.35 & 14.25 & 31.84 & 1.42 & 8.53 & 18.76\\
    \bottomrule
    \end{tabular}
\end{table}

%% file: tables/ablation_vse.tex
\begin{table}[t!]
    \small
    \centering
    \tabcolsep 4pt
    \caption{Ablation study on Cross-modal Transformer}
    \label{tab:ablation_vse}
    \centering
    \vspace{0.1cm}
    \begin{tabular}{l@{}c ccc@{\quad\;\;}ccc}
    \toprule
    \multirow{2}{*}{Model} & \multirow{2}{*}{\makecell[c]{X-Modal}} & \multicolumn{3}{c}{IoU=0.5} & \multicolumn{3}{c}{IoU=0.7} \\
    \cmidrule(lr){3-5} \cmidrule(lr){6-8}
    & & R1 & R10 & R100 & R1 & R10 & R100 \\
    \midrule
    \multirow{2}{*}{\ourmodel} & \xmark & 1.38 & 8.89 & 26.35 & 0.84 & 5.08 & 15.27 \\
     & \cmark & \textbf{2.94}  & \textbf{14.49}  & \textbf{32.49}  & \textbf{1.74} & \textbf{8.75} & \textbf{19.08} \\
    \bottomrule
    \end{tabular}
\end{table}

%% file: tables/ablation_clip_len.tex
\begin{table}[t!]
    \small
    \centering
    \tabcolsep 4pt
    \caption{Ablation study on different clip lengths}
    \vspace{0.1cm}
    \label{tab:ablation_clip_len}
    \centering
    \begin{tabular}{l@{}c ccc@{\quad\;\;}ccc}
    \toprule
    \multirow{2}{*}{Model} & \multirow{2}{*}{\makecell[c]{Clip \\ Length}} & \multicolumn{3}{c}{ IoU=0.5 } & \multicolumn{3}{c}{ IoU=0.7 } \\
    \cmidrule(lr){3-5} \cmidrule(lr){6-8}
    & & R1 & R10 & R100 & R1 & R10 & R100 \\
    \midrule
    \multirow{3}{*}{\ourmodel} & 16 & 2.70 & 14.06 & 31.85 & 1.63 & 8.16 & 18.60 \\
     & 32 & \textbf{2.94}  & {14.49}  & \textbf{32.49}  & \textbf{1.74} & {8.75} & \textbf{19.08} \\
     & 64 & 2.78 & \textbf{14.69} & 32.08 & 1.70 & \textbf{9.00} & 18.71 \\
    \bottomrule
    \end{tabular}
\end{table}

%% file: tables/ablation_weightshare.tex
\begin{table}[t]
    \small
    \centering
    \tabcolsep 4pt
    \caption{Ablation study on weight sharing}
    \label{tab:ablation_weightshare}
    \vspace{0.1cm}
    \begin{tabular}{l@{}c ccc@{\quad\;\;}ccc}
    \toprule
    \multirow{2}{*}{Model} & \multirow{2}{*}{\makecell[c]{Weight \\ Sharing}} & \multicolumn{3}{c}{IoU=0.5} & \multicolumn{3}{c}{IoU=0.7} \\
    \cmidrule(lr){3-5} \cmidrule(lr){6-8}
    & & R1 & R10 & R100 & R1 & R10 & R100 \\
    \midrule
    \multirow{2}{*}{\ourmodel} & \xmark & \textbf{2.94}  & \textbf{14.49}  & \textbf{32.49}  & \textbf{1.74} & \textbf{8.75} & \textbf{19.08}  \\
     & \cmark & 2.89 & 14.17 & 30.31 & 1.69 & 8.05 & 17.24 \\
    \bottomrule
    \end{tabular}
\end{table}

%% file: tables/ablation_clippos.tex
\begin{table}[t!]
    \small
    \centering
    \tabcolsep 4pt
    \caption{Ablation study on clip position embeddings}
    \label{tab:ablation_clippos}
    \centering
    \vspace{0.1cm}
    \begin{tabular}{l@{}c ccc@{\quad\;\;}ccc}
    \toprule
    \multirow{2}{*}{Model} & \multirow{2}{*}{\makecell[c]{Clip \\ Position}} & \multicolumn{3}{c}{IoU=0.5} & \multicolumn{3}{c}{IoU=0.7} \\
    \cmidrule(lr){3-5} \cmidrule(lr){6-8}
    & & R1 & R10 & R100 & R1 & R10 & R100 \\
    \midrule
    \multirow{2}{*}{\ourmodel} & \xmark & 2.82 & 14.39 & 32.01 & 1.76 & 8.59 & 18.63 \\
     & \cmark & \textbf{2.94}  & \textbf{14.49}  & \textbf{32.49}  & \textbf{1.74} & \textbf{8.75} & \textbf{19.08} \\
    \bottomrule
    \end{tabular}
\end{table}

%% file: 5_conclusion.tex
% !TEX root = main_cvpr21.tex
\section{Conclusion}
\label{sec:conclusion}

In this paper, we propose a hierarchical multi-modal encoder ({\ourmodel}) that captures video dynamics in three scales of granularity, frame, clip, and video. By hierarchically modeling videos, {\ourmodel} achieves significantly better performance than the baseline approaches on moment localization task in video corpus, and further establishes new state-of-the-art on two challenging datasets, ActivityNet captions and TVR. Extensive studies verify the effectiveness of the proposed architectures and learning objectives.

%% file: supp_content.tex
\subsection{Additional Implementation Details}
\label{supp:impl}

\begin{figure*}[tbh]
    \centering
    \small
    \includegraphics[width=\textwidth]{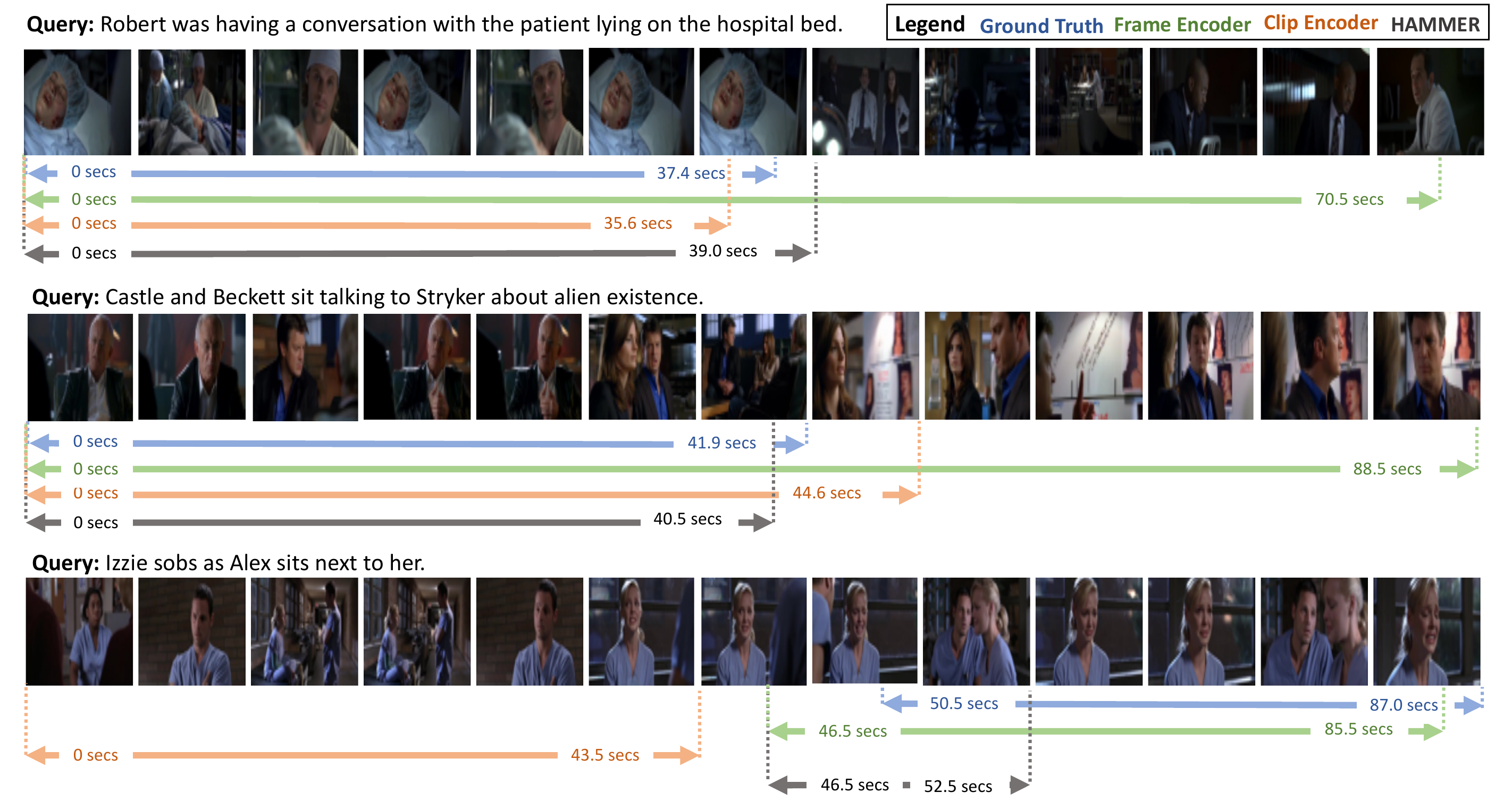}
    \vspace{-10pt}
    \caption{Illustration of temporal localization results using {\ourmodel} and its individual hierarchies, the frame and clip encoder. The top two are successful examples and the bottom one is a failed example.}
    \label{fig:qualitative_result}
\end{figure*}

\subsubsection{Visual Feature Representation}

We use three different kinds of visual features throughout our experiments, \ie, ResNet-152~\cite{he2016deep}, I3D~\cite{carreira2017quo}, and their combination.
On the ActivityNet Captions dataset, we report model performance with ResNet-152 for fair comparison with prior methods, and also report results using the widely used I3D features for comparison.
On the TVR dataset, we follow the setting in~\cite{lei2020tvr} and report results using the concatenation of ResNet-152 and I3D features. The details about how to extract these features are specified below.

\mypara{ResNet-152 Feature} For all of the frames in a given video, we extract features ($2048$ dimensional) from the penultimate layer of a ResNet-152~\cite{he2016deep} model pre-trained on the ImageNet dataset. For ActivityNet Captions dataset, the frames are extracted at a rate of $6$ FPS.

\mypara{I3D Feature} We use an I3D model~\cite{carreira2017quo} to extract the spatio-temporal visual features (with a dimension of 1024). The I3D model used for feature extraction is pre-trained on the Kinetics-400~\cite{kay2017kinetics} dataset. Similar to the setting of ResNet-152 features, we take the features from the penultimate layer of the I3D model. For ActivityNet Captions dataset, the I3D features are extracted at a frame rate of $1$ FPS.

\mypara{I3D$+$ResNet-152 Feature}
For the TVR dataset, we use the I3D$+$ResNet-152 features provided by Lei~\etal~\cite{lei2020tvr} to represent the visual information in the videos.
The I3D and ResNet-152 models are pre-trained on Kinetics-600 \cite{carreira2018short} and ImageNet datasets, respectively. For both models, the features from the penultimate layers are used. The ResNet-152 features are extracted at a rate of $3$ FPS and max-pooled over each $1.5$-seconds clip. The I3D features are extracted for every $1.5$ seconds as well. The two sets of features are then concatenated to form the combined $3072$-dimensional feature.

\subsubsection{Subtitle (ASR) Feature Representation}
Previous work~\cite{lei2020tvr} have demonstrated that subtitles (e.g., extracted from ASR) can complement the visual information in video and language tasks. For the TVR dataset, we follow the standard setting and use the pre-extracted ASR embeddings provided by Lei~\etal~\cite{lei2020tvr} as an additional input to our models. Contextualized token-level subtitle embeddings are first generated using a 12-layer RoBERTa~\cite{Liu2019roberta} model fine-tuned on the TVR train split. The token embeddings are then max-pooled every 1.5 seconds to get an aggregated $768$-dimensional feature vector. A zero vector of the same dimensionality is used for frames without corresponding subtitles. The resulting subtitle embeddings are temporally aligned to the visual features (I3D$+$ResNet-152), allowing us to combine the two modalities later in the cross-modal encoders.
We refer the reader to Lei~\etal~\cite{lei2020tvr} for more details on the feature extraction process.

\subsubsection{Model Architecture with ASR input}
The general architecture of the \ourmodel model is illustrated in Figure 1 of the main text. It consists of 2 hierarchical encoders (\ie, the frame and clip encoders) that have the same structure, and two input streams, query and video. When only the visual features are present (\eg, ActivityNet Captions), the video encoder contains only the visual encoder. Each hierarchical encoder contains 5 standard Transformer layers for the query input and 1 Transformer layer for the visual input. There is an additional cross-modal Transformer layer between the query and visual representations.

When ASR is provided as another input stream (\eg, in TVR), we add another branch to the video encoder in each of the hierarchical encoders to process the ASR input. The ASR and visual branches have similar structure as both have 1 Transformer layer. The pre-extracted ASR embeddings and the visual features both attend to the query representation to form their contextualized representations. The query embeddings in turn attend to the ASR and visual modalities separately. The resulting ASR- and visual-grounded query representations are then added together in the feature dimension, followed by a normalization and a dropout layer. Finally, the query-grounded ASR and visual representations are concatenated to form the frame-level and clip-level representations for the two hierarchical encoders, respectively.

\subsubsection{Model Optimization}

For ActivityNet Captions, we train the models with a mini-batch size of 64 and optimize them using Adam~\cite{kingma2014adam} with a maximum learning rate of \num{4e-5}. The learning rate increases linearly from 0 to the max rate in the first 10\% training epochs and then drops to $0.1\times$ and $0.01\times$  of the max rate at 50\% and 75\% of the training epochs, respectively. We set the maximum video sequence length to be 128, and experiment with clip lengths varying from 16 to 64 (refer to Table 7 in the main text).

For TVR data, we train the model with a batch size of 128. We use the same learning rate schedule as mentioned above but with a maximum learning rate of \num{2e-4}. We set the maximum video sequence length to 96, and experiment with clip lengths varying from 8 to 48.

\subsubsection{Model Initialization}

We do not pre-train \ourmodel on any dataset. We randomly initialize the visual and ASR branches and the cross-attention Transformer layers. For the text query branch, following prior work~\cite{lu2019vilbert}, we initialize from the first 5 layers of a pre-trained BERT model~\cite{devlin2018bert}, and use the uncased WordPiece tokenizer~\cite{wu2016google} to tokenize the text input. The vocabulary size of the tokenizer is 30,522.

\subsection{Illustration on the TVR Dataset}
\label{supp:viz}

Figure \ref{fig:qualitative_result} illustrates the temporal localization results on TVR dataset using predictions from {\ourmodel} and its frame and clip encoders. In the top 2 examples, {\ourmodel} successfully localizes the video segments described by the respective queries with the help of the clip encoder even though the frame encoder makes erroneous predictions. In the bottom-most example, the clip encoder picks the incorrect video clip, causing {\ourmodel} to only partially capture the video segment described by the query. These examples show the important role played by the two hierarchical encoders---while the clip encoder is responsible for choosing the video clips that best describe the query, the frame encoder fine-tunes the predictions within the chosen clips.